# EMOTION DETECTION AND MUSIC RECOMMENDATION SYSTEM

Swetha Kambham [1], Hubert A, Johnson [1] and Sai Prathap Reddy Kambham [2]

[1] Department of Computer Science, Montclair State University, Montclair, NJ 07043
[2] Full Stack Developer, ERP Health LLC, Philadelphia, PA 19106

## ABSTRACT

*As artificial intelligence becomes more and more ingrained in daily life, we present a novel system that uses deep learning for music recommendation and emotion-based detection. Through the use of facial recognition and the DeepFace framework, our method analyses human emotions in real-time and then plays music that reflects the mood it has discovered. The system uses a webcam to take pictures, analyses the most common facial expression, and then pulls a playlist from local storage that corresponds to the mood it has detected. An engaging and customised experience is ensured by allowing users to manually change the song selection via a dropdown menu or navigation buttons. By continuously looping over the playlist, the technology guarantees continuity. The objective of our system is to improve emotional well-being through music therapy by offering a responsive and automated music-selection experience.*

## 1. INTRODUCTION

Facial emotion detection is an important area of study in computer vision and artificial intelligence because of its potential far-reaching application in education, cybersecurity, Immigration and Border Protection Control, and general commercial usage.

Facial expressions are employed in nonverbal communication, to identify individuals, and to help identify emotions. They are just as significant as voice tone in everyday emotional communication [20]. Additionally, they serve as a feeling indicator, enabling a man or woman to communicate an emotional state [21][25]. People are able to identify someone's emotional condition right away. As a result, automatic emotion recognition algorithms frequently incorporate facial expression data [17].

The effort to develop this Emotion Detection and Music Recommendation System was driven by the desire to automate the process of detecting an individual's emotion and alleviate the challenges that music lovers face in the manual selection of music based on their emotional state/mood. The automatic selection of music is aimed at providing an upbeat companion-like experience/feeling for the listener.

There are many unique experiences that users of this system will gain from its use. For example, a user who may be feeling 'down' may feel comforted and inspired and realize that he/she is not alone. The emotion detection and music recommendation system focus on detecting and catering to the following emotions depicted in Figure 1.

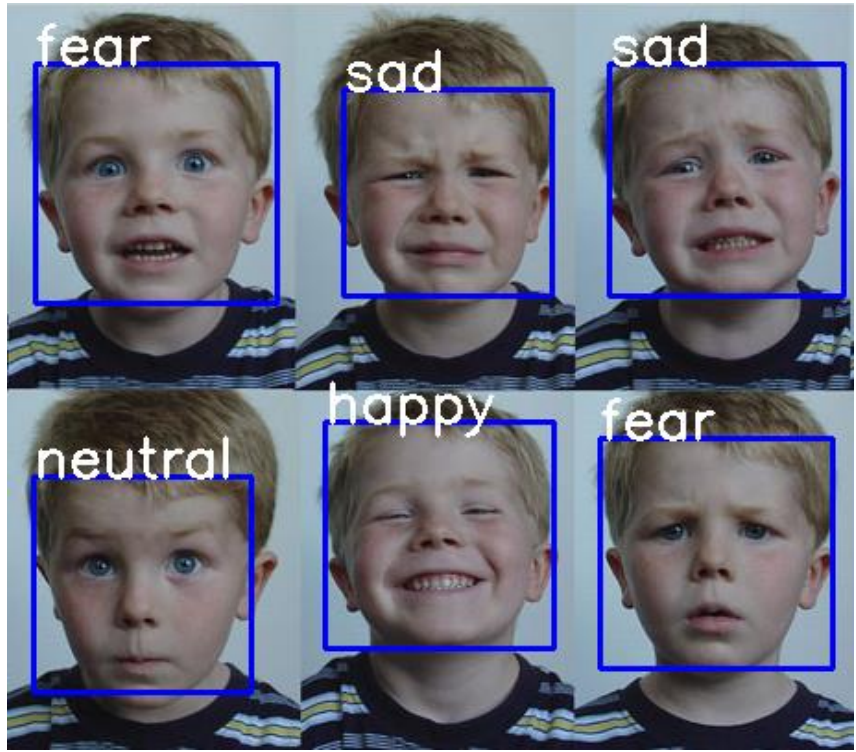

Figure 1. Emotion Expressions

Once a particular emotion is detected, music deemed befitting the emotion will automatically be selected and played from an existing playlist. The playlists utilized in this endeavor are illustrated in Figure 2. Bearing in mind that emotions transition from one stage/phase to another, so too the system allows the user to override the automatic music selection choice and make a selection from a different playlist to suit their mood.

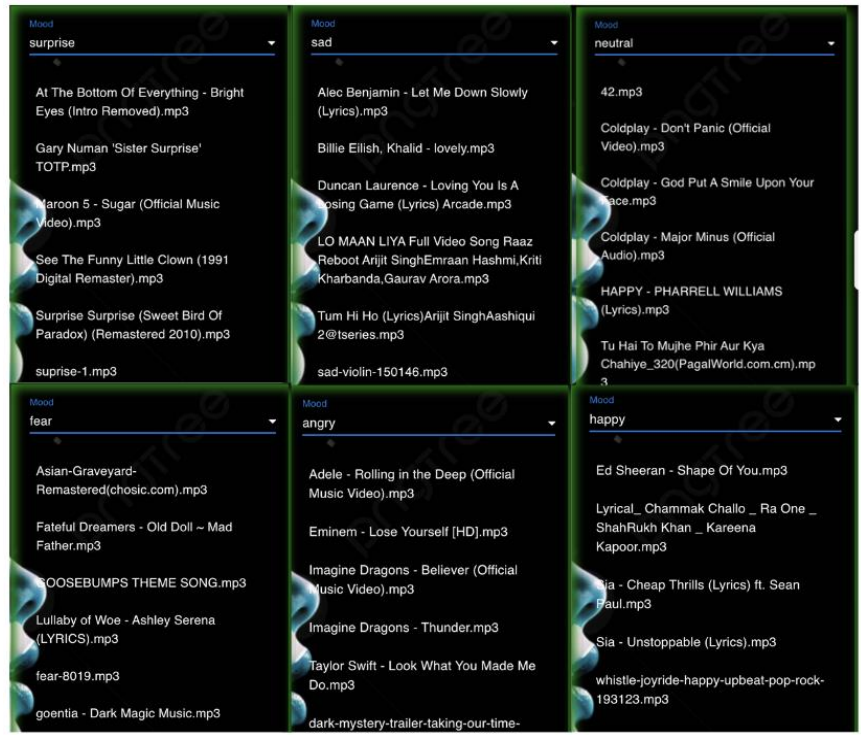

Figure 2. Mood Detection and Music Recommendation Playlist

Visual/Facial expressions are means by which individuals communicate to others their current emotional state, Facial expressions are important indicators of an individual's current emotional state [9]. According to several surveys [18] [13], verbal components convey one-third of human communication and while non-verbal components convey two-thirds. Facial expressions are one of the primary information channels in interpersonal communication among the various non-verbal cues. It is little wonder, therefore, that much attention has been directed at researching/studying facial emotions. Studies have also shown that the process of interpreting facial expressions is subject to cultural and individual differences. These nuances were, however, not considered in this paper.

Currently, there is a set of principles which highlight distinctly different types of emotions and their roles in the evolution and development of different levels of consciousness/awareness and of mind, human mentality, and behavior. The ongoing reformulations of these principles are facilitated by advances in emotion science, cognitive neuroscience, and developmental clinical science, as well as in social and personality psychology.

They, among others, led to a new perspective on emotion-related theoretical development and research on emerging concepts such as the impact of music on the emotional experiences, empathy, and sympathy and their relations to emotion schemas.

The purpose of this system is to (1) detect the mood of the user and determine the user's emotion, and (2) automatically select and play a song/music which is commensurate with the detected emotion. In this system, Although the default music will align with the detected mood, the user will have the option to choose a different song/music. For example, a user whose detected emotion is 'sad' may prefer to listen to an up-beat rhythm to get himself/herself out of the current sad state.

## 2. Related Work

Emotion-based recommendation systems are designed to enhance user satisfaction by incorporating emotional data into the recommendation process[21].Emotion recognition using facial expressions is a rapidly expanding science with important implications in virtual reality, human-computer interface (HCI), and health monitoring. This research topic focuses on interpreting human emotions via facial clues, utilising various technology breakthroughs to increase accuracy and application.

Facial expression analysis in real-world situations presents substantial hurdles, such as lighting changes, major position fluctuations, and partial or complete occlusion. The process of identifying emotions is made more difficult by these elements, which have an impact on the completeness and sharpness of facial features [15][28]. Some researchers have proposed new architectures and methodologies, such as meta-learning and landmark detection, to improve emotion recognition in the presence of occlusions[14].

Applications needing high precision in facial recognition tasks are better suited for VGG-Face models [4] since they generally offer superior feature extraction capabilities and higher accuracy than DeepFace models [1].

The field of emotion recognition is one that is developing quickly and is essential to improving healthcare, human-computer interface (HCI),Music Therapy, and other businesses. Emotion recognition technology, by allowing systems to comprehend and respond to human emotions, has the potential to greatly improve user experience and interactions. Emotion recognition in HCI improves user pleasure and engagement by enabling more responsive and intuitive interfaces [5][8].

## 3. Types of Emotions and Importance of music

Emotions are usefully divided into two broad types or kinds– basic emotion episodes and dynamic emotion-cognition interactions or emotion schemas.Basic emotion refers to any emotion that is assumed to be fundamental to human mentality and adaptive behavior.

In this category there is basic positive emotion – joy and happiness, and Basic negative emotions (sadness, anger, disgust, fear) The intent of the Emotion Detection and Music Recommendation System is to help the user get the companion for positive emotions and negative emotions.

The system gives the user an opportunity to change the automatically-selected song/music by providing a dropdown menu to change the playlist and can list songs from which he/she can select the song from the list or can click on the "next" button which he/she desires.

### 3.1. Why do People Listen to Music?

It is theorized that there are three distinct underlying reasons people listen to music: (1) People listen to music to regulate arousal and mood, (2) to achieve self-awareness, and (3) as an expression of social relatedness. For this paper, the first two reasons were judged to be of greater interest than the third—the idea that music has evolved primarily as a means for social cohesion and communication. Listening to music is one of the most enigmatic of human behaviors. In the array of seemingly odd behaviors, few behaviors match music for commandeering so much time, energy, and money. Listening to music is also one of the most popular leisure activities, and it is a constant companion to people in their everyday lives.

Another line of theorizing refers to music as a means of social and emotional communication. For example, Panksepp and Bernatzky (2002, p. 139) [19] [23] argued that in social creatures like ourselves, whose ancestors lived in arboreal environments where sound was one of the most effective ways to coordinate cohesive group activities, reinforce social bonds, resolve animosities, and to establish stable hierarchies of submission and dominance, there could have been a premium on being able to communicate shades of emotional meaning by the melodic character (prosody) of emitted sounds.

- The most common motive for listening to music is to influence emotions.
- Music helps to channel one's frustration or purge (catharsis) negative emotions.

Music has the capability of evoking profound emotions (chills and thrills) in listeners. A piece of music is often referred to as sad, joyful, tender, or harsh. The following describes key factors that explain our emotional responses to music, according to Juslin, 2019 [12].

(1)Brain stem response: The brainstem response is an evolutionary response to any kind of sound that evokes arousal. The brain stem response to sound explains why music in general is pleasurable.(2) Rhythmic Entrainment: Rhythmic entertainment refers to the way the listener moves in synchrony to the beat (for example, dancing, head movement, marching, or foot tapping). Moving in sync with music gives a sense of pleasure. It is no coincidence that rhythmic dance-like music makes people happy because it is easy to entrain (attune) to its rhythmic pattern [2].(3)Emotional contagion: Music doesn't only evoke emotions at the individual level, but also at the interpersonal level. When people attend a dance party or a concert, their emotions are influenced not just by the music being played, but also in part by the emotions exhibited by other people present at these events. (4)Musical expectancy: Studies have shown that it is the creation of expectations that makes music so emotionally powerful. Research shows that anticipation is a key element in activating the reward system and provoking musical pleasure. Unexpected change in musical features intensity and tempo is one of the primary means by which music provokes a strong emotional response in listeners (Salimpoor and colleagues, 2015) [22].(5)Pleasurable sadness: According to Huron, 2011 [11], Sad music evokes a special emotion that people enjoy listening to. Listening to sad music can help one to get rid of a negative emotion, without actually experiencing the loss. Listening to a sad song while in a sad mood can potentially have the same effect as someone (a friend) empathizing with the listener's experience. It can bring comfort by making one feel connected and a sense of not being so alone.(6)Aesthetic emotion: Music is one of the most powerful means of evoking aesthetic emotions. Aesthetic emotions include feeling moved, awe [2], wonder, transcendence, nostalgia, and tenderness. In response to these emotions, we may experience goosebumps, and motivation for the improvement of self and society (Winner, 2019).

One of the most important issues in the psychology of music is how music affects emotional experience (Juslin, 2019) [12]. Music has the ability to evoke powerful emotional responses such as chills and thrills in listeners.

Positive emotions dominate musical experiences. Pleasurable music may lead to the release of neurotransmitters associated with reward, such as dopamine [7]. Listening to music is an easy way to alter mood or relieve stress [24]. People use music in their everyday lives to regulate, enhance, and diminish undesirable emotional states (e.g., stress, fatigue). How does music listening produce emotions and pleasure in listeners?.

In sum, music is capable of inspiring emotions. Music can be used to create an emotional atmosphere such as calming, relaxing, playful, sincere, or intimate. These factors also explain an important emotional benefit of music for mood regulation.

## 4. METHODOLOGIES

The system combines Haar Cascade, DeepFace and Howler.js to identify faces, assess emotions, and play music that corresponds to the mood it detects.

**Haar Cascade for Face Detection**

A machine learning-based technique for object detection, Haar Cascade is used in computer vision tasks, especially face detection. It was developed by Viola and Jones in 2001 and is widely used for real-time face detection [26]. It follows these steps for face detection:

- The Haar Cascade classifier (haarcascade_frontalface_default.xml) from OpenCV is loaded for face detection.
- The input image, converted from base64 format, is processed using OpenCV's detect MultiScale() method to identify faces.
- The method returns face coordinates, which are used to extract the region of interest (ROI) for further processing, such as emotion detection.

**DeepFace for Emotion Detection**

A deep learning-based library for facial recognition and emotion analysis, DeepFace uses pre-trained models to identify emotions in facial expressions [6] [27].It follows these steps for face detection:

- Initially, the system finds a face in the input picture. Using the parameter actions=['emotion'], the Region of Interest (ROI) that contains the face is extracted and supplied to DeepFace's analyze() method.
- DeepFace determines the dominant emotion (such as happiness, sadness, anger, etc.) by analyzing the facial expression. The system is able to identify the most likely emotional state of the person in the picture by using the confidence scores it returns for each emotion.

**Howler.js for Audio Playback**

Howler.js is a JavaScript library that enables seamless audio playback and control within web applications. It makes sure that it works with various browsers and offers a simple interface for audio control [10]. It follows these steps for face detection:

- The system chooses a suitable audio track that corresponds to the emotional state based on the emotion that was detected (playing a lively song for a "happy" emotion, for example).
- The playback of the chosen audio is handled by Howler.js, which also controls playback rate, looping, and volume control. This enables the system to deliver dynamic and fluid audio experiences that are adapted to the emotional state of the user.

The proposed development of the Emotion Detection and Music Recommendation system follows a structured flow of control, incorporating facial emotion detection, emotion-based song selection, and an interactive user interface. The methodology consists of the following key steps:

- **Image Capture:** The system activates the webcam and captures a user's image every three seconds, ensuring real-time analysis.
- **Emotion Detection:** The captured image is processed using the Haarcascade Frontal

Face classifier to detect facial regions. The DeepFace library then analyzes the face and determines the dominant emotion based on probability values.
- **Emotion-Based Song Selection:** The detected emotion is mapped to a predefined set of moods (e.g., happy, sad, angry, calm). A corresponding playlist is retrieved from local storage, and a random song from the selected playlist is played.
- **User Interaction and Control:** Users can manually change the song using a dropdown menu or navigation buttons (previous/next). The playlist operates in a loop, ensuring continuous playback.

What differentiates this system from other existing emotion-based recommendation systems is that with this system, it is using deepface library which has models pretrained which gives accuracy of 97.5% along with that the time-lapse between detected mood and the automatic playing of the chosen mood-appropriate music/song is minimal. Table 1 below shows the times taken to detect an emotion and the time to retrieve and play a song/music. The audio is activated and played instantaneously upon detection of an emotion. In addition, this system offers the user the option of making a different selection of a song/music from a playlist, if he/she feels the need to choose a song different from the one which is automatically chosen by the system.
.

Table 1: Time to Detect Emotion and Time to Play Related Song

|  | Time to Detect Emotion | Time for Song Retrieval |
| --- | --- | --- |
| Minimum | 0.019 milliseconds | 0.028 milliseconds |
| Maximum | 0.043 milliseconds | 0.531 milliseconds |
| Average | 0.031 milliseconds | 0.279 lliseconds |

## 4.1. System Architecture and Flow

The architecture of the proposed system, as depicted in Figure 3, follows a structured workflow that ensures seamless interaction between the components:

- **System Initialization:** The application launches, activating the webcam and initializing the DeepFace model.
- **Image Processing and Emotion Detection:** The system captures images from the webcam and uses Haarcascade Frontal Face detection to identify facial regions. DeepFace then analyzes the image and determines the dominant emotion.
- **Music Recommendation Engine:** The system maps the detected emotion to a predefined playlist stored locally. A random song from the selected playlist is played.
- **User Interaction:** Users can manually change songs via a dropdown menu or use navigation buttons (previous/next) to control playback.
- **Looping Mechanism:** Playlists operate in a continuous loop, ensuring uninterrupted music playback.

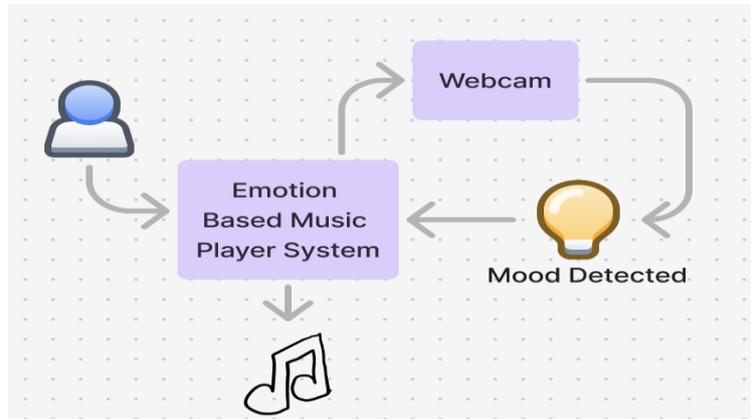

Figure 3. Emotion Detection and Music Recommendation Architecture

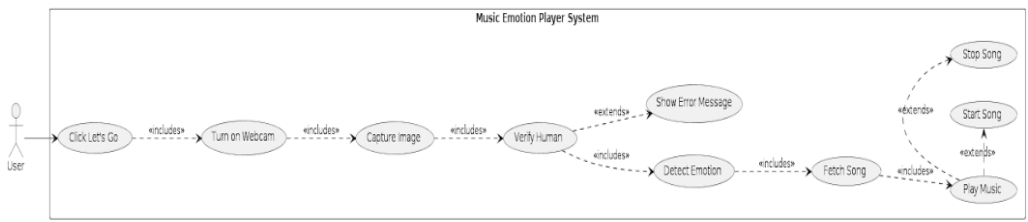

Figure 4. Emotion Detection Use Case Diagram

## 5. ANALYSIS

Conducted an analysis on a dataset of 34 images, each image containing the group of 2 or 3 individuals, to examine the dominant and most frequently occurring emotions detected by DeepFace [6]. The following observations were made:

- **Highest Percentage Analysis**: Each image was analyzed individually to determine the dominant emotions with the highest percentage as shown in Figure 5. The results showed that neutral and happy were the most frequently identified emotions across the dataset.

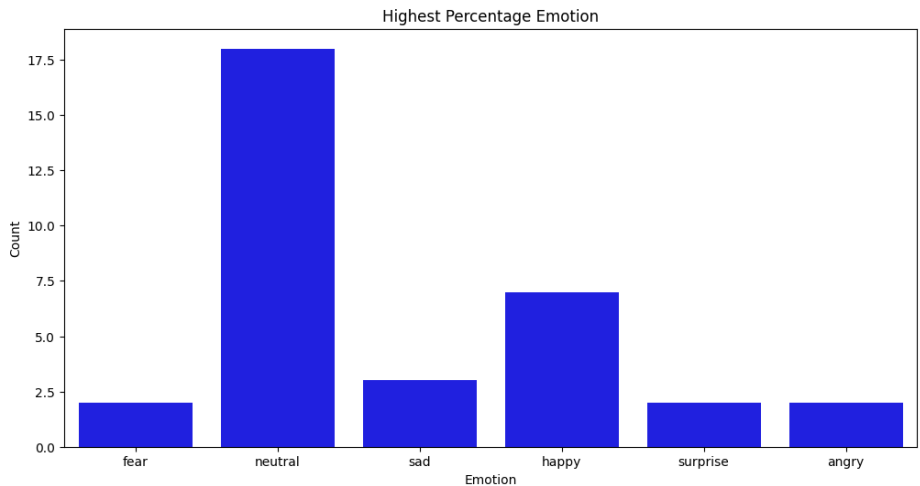

Figure 5. Emotion with Highest percentage

- **Most Frequent Analysis:** By considering the most frequently occurring emotions across multiple images, we found that angry and sad emotions appeared most often, as shown in Figure 6, suggesting that these emotions are more consistently detected across different individuals.

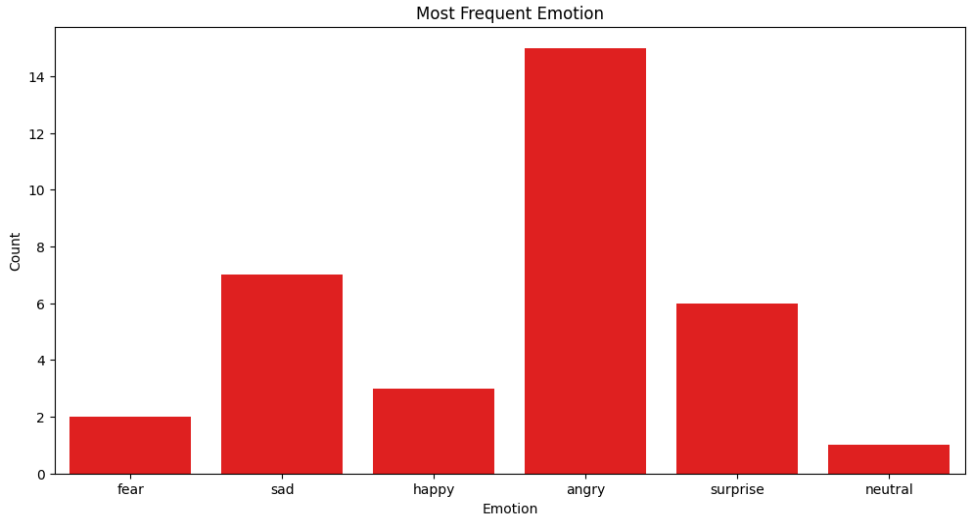

Figure 6. Most Frequent Emotion

- **Comparison of both approaches**: By comparing both approaches, we observe that the analysis based on the highest percentage of detected emotions shows a greater overall density than the one based on the most frequently occurring emotions. In the highest percentage analysis, neutral is the dominant emotion, whereas in the most frequent analysis, angry appears most often. This distinction highlights how different analytical methods can influence the interpretation of emotion detection results, as shown in Figure 7.

    ○ The "highest percentage" approach determines the emotion with the strongest confidence score in an image. For example, if an image contains three people—one expressing happiness at 90% confidence and two expressing sadness at 60%—this method would classify the image as happy because of the highest confidence score.
    ○ The "Most frequent emotion" approach considers the emotion that appears the most across the detected faces. In the same scenario, this method would classify the image as sad because sadness appears more frequently among the individuals.

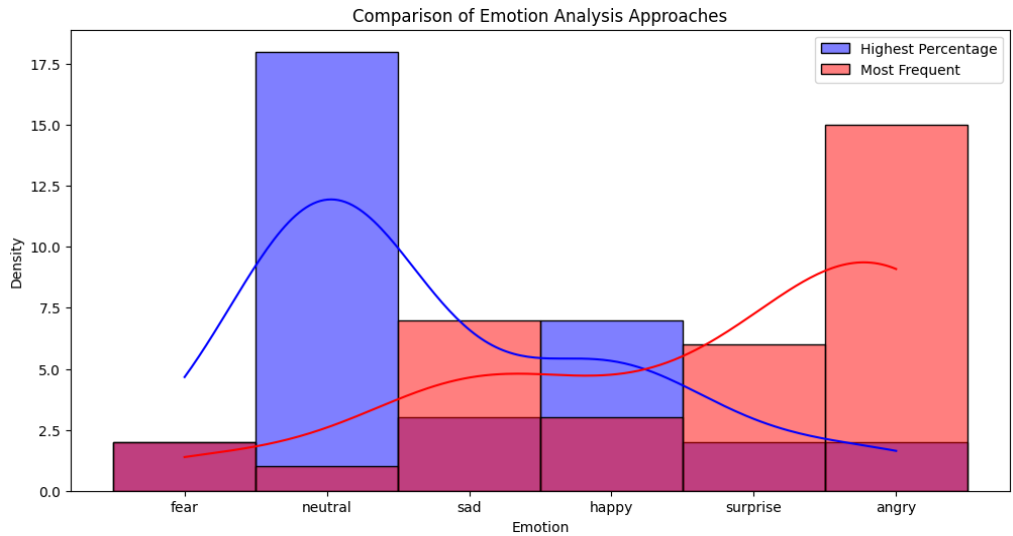

Figure 7. Highest Percentage and Most frequent emotion comparison

## 6. CHALLENGES

Although the use of DeepFace for real-time facial recognition was remarkably accurate under varied conditions tested, the system performance under low lighting conditions was not very good. but its accuracy in detecting emotions of a subject in complete, or almost complete darkness was not examined. The system was also not able to accurately detect emotions of facial images captured at an angle; for improved accuracy, the subject whose facial image is being captured needs to be directly in front of the camera. For best results, in terms of accurately capturing the emotion, the subject needs to be no more than 2 feet 6 inches from the camera. The major challenges are to determine the accuracy of the system in detecting emotions when:

- The subject's face is partially covered, for example with a mask, or while wearing sunglasses.
- Using the system installed on a range of devices with varying degrees of camera-resolutions
- The subject's face is in complete darkness
- Improving the accuracy when the detection is done from an angle

Because the facial image captured by the current system is not stored, and the system is (or will be) deployed on a personal device (a Smartphone or a personal computer) the threat to personal data security and ethical concerns are non-existent or minimal.

## 7. POTENTIAL ENHANCEMENTS AND APPLICATIONS

A potential enhancement to the system which the authors envision, is to provide a feature for users to provide feedback, which could potentially lead to enhancing current features and/or incorporating additional functionalities. One such feedback which is envisioned is that of giving users the ability to indicate the genre of music they prefer. Such feedback when compared with the developers' playlist could lead to providing a wider playlist selection and thereby being able to reach a wider audience. Future enhancements will undoubtedly bring ethical concerns and privacy issues into question if the system is expanded and enhanced for commercial use. Further Potential Applications of this System Include:

- Its use, voluntarily by individuals or at the recommendation of a physician, or a therapist for stress management.
- Being an impactful agent in a work environment for certain occupations, a motivating vehicle on workers' productivity.
- Its utility for commercial uses, for example to let patrons feel welcome in a building or store, and to monitor patrons in a store.
- Used for security purposes, example to monitor patrons entering or leaving a high-security environment
- Taking attendance in a class or at a social event
- Integration with smart wearable devices and smart-home systems.

## 8. ACKNOWLEDGEMENTS

The authors would like to extend special thanks to the Software Engineering students, and the volunteers who experimented with the Emotion Detection and Recommendation System and provided invaluable feedback during the development of the system. Thanks is also due to the authors whose resources were consulted during the production of this system.

## 9. CONCLUSION

This study presents an AI-driven emotion-based music recommendation system leveraging DeepFace for real-time facial emotion detection. By analyzing facial expressions, the system dynamically selects and plays music to match or uplift the detected emotional state. Our analysis demonstrated that certain emotions, such as happy, sad, neutral angry are more frequently detected as dominant emotions. The ability to override automatic selection enhances user experience, allowing for emotional regulation through music. Future work includes integrating an API for song streaming instead of local storage, expanding the system to mobile platforms, and implementing user authentication for personalized experiences.

## REFERENCES

[1] Awana, S. V., Singh, A., Mishra, A., Bhutani, V., Kumar, S. R., & Shrivastava, P. (2023). Live Emotion Detection Using Deepface. *6th International Conference on Contemporary Computing and Informatics (IC3I)*, Gautam Buddha Nagar, India, 581-584. doi: 10.1109/IC3I59117.2023.10
[2] awe, https://www.psychologytoday.com/us/basics/awe
[3] binaural beats, https://www.psychologytoday.com/us/basics/binaural-beats
[4] Chen, H., & Haoyu, C. (2019). Face Recognition Algorithm Based on VGG Network Model and SVM. *Journal of Physics: Conference Series*, 1229. https://doi.org/10.1088/1742-6596/1229/1/012015.
[5] Cowie, R., Douglas-Cowie, E., Tsapatsoulis, N., Votsis, G., Kollias, S., Fellenz, W., & Taylor, J. (2001). Emotion recognition in human-computer interaction. *IEEE Signal Process. Mag.*, 18, 32-80. https://doi.org/10.1109/79.911197.
[6] Deepface Documentation, https://github.com/serengil/deepface
[7] dopamine, https://www.psychologytoday.com/us/basics/dopamine
[8] Fragopanagos, N., & Taylor, J. (2005). Emotion recognition in human-computer interaction. *Neural networks: the official journal of the International Neural Network Society*, 18 4, 389-405. https://doi.org/10.1016/j.neunet.2005.03.006.
[9] Horstmann, G. (2003). What do facial expressions convey: feeling states, behavioral intentions, or action requests?. Emotion, 3 2, 150-66 . https://doi.org/10.1037/1528-3542.3.2.150.
[10] Howler.js Documentation, https://github.com/goldfire/howler.js#documentation
[11] Huron, David. (2011). Why is sad music pleasurable? A possible role for prolactin. Musicae Scientiae. 15. 146-158. 10.1177/1029864911401171.

[12]  Juslin, Patrik. (2019). Musical Emotions Explained: Unlocking the Secrets of Musical AffectUnlocking the Secrets of Musical Affect. 10.1093/oso/9780198753421.001.0001.

[13]  Kaulard K., Cunningham D.W., Bülthoff H.H., Wallraven C. The MPI facial expression database—A validated database of emotional and conversational facial expressions. *PLoS ONE*. 2012;7:e32321. doi: 10.1371/journal.pone.0032321.

[14]  Kuruvayil, S., & Palaniswamy, S. (2021). Emotion recognition from facial images with simultaneous occlusion, pose and illumination variations using meta-learning. *J. King Saud Univ. Comput. Inf. Sci.*, 34, 7271-7282. https://doi.org/10.1016/j.jksuci.2021.06.012.

[15]  Lian, Z., Li, Y., Tao, J., Huang, J., & Niu, M. (2019). Expression Analysis Based on Face Regions in Real-world Conditions. *International Journal of Automation and Computing*, 17, 96 - 107. https://doi.org/10.1007/s11633-019-1176-9.

[16]  Liu, Z., Xu, W., Zhang, W., & Jiang, Q. (2023). An emotion-based personalized music recommendation framework for emotion improvement. Inf. Process. Manag., 60, 103256. https://doi.org/10.1016/j.ipm.2022.103256.

[17]  Mao Q., Pan X., Zhan Y., Shen X., Using Kinect for real-time emotion recognition via facial expressions, Frontiers Inf Technol Electronic Eng, 16 (2015), no. 4, 272–282. [5] Li B. Y. L., Mian A. S., Liu W., Krishna A., Using Kinect for face recognition under varying poses, expressions, illumination and disguise, 2013 IEEE Workshop on Applications of Computer Vision (WACV), 2013, 186–192.

[18]  Mehrabian A. Communication without words. *Psychol. Today.* 1968;2:53–56.

[19]  Panksepp, Jaak & Bernatzky, Günther. (2002). 'Emotional Sounds and the Brain: the Neuro-affective Foundations of Musical Appreciation'. Behavioural processes. 60. 133-155. 10.1016/S0376-6357(02)00080-3.

[20]  Przybyło J., Automatyczne rozpoznawanie elementów mimiki twarzy w obrazie i analiza ich przydatności do sterowania, rozprawa doktorska, Akademia Górniczo-Hutnicza, Kraków, 2008.

[21]  Ratliff M. S., Patterson E., Emotion recognition using facial expressions with active appearance modelCTA Press, Anaheim, CA, USA, 2008, 138–143

[22]  Salimpoor VN, Zald DH, Zatorre RJ, Dagher A, McIntosh AR. Predictions and the brain: how musical sounds become rewarding. Trends Cogn Sci. 2015 Feb;19(2):86-91. doi: 10.1016/j.tics.2014.12.001. Epub 2014 Dec 19. PMID: 25534332

[23]  Schäfer T, Sedlmeier P, Städtler C, Huron D. The psychological functions of music listening. Front Psychol. 2013;4:511. Published 2013 Aug 13. doi:10.3389/fpsyg.2013.00511

[24]  stress, https://www.psychologytoday.com/us/basics/stress

[25]  Tian s, Proceedings of the Third IASTED International Conference on Human Computer Interaction, AY. I., Kanade T., Cohn J. F., Recognizing action units for facial expression analysis, IEEE Transactions on Pattern Analysis and Machine Intelligence, 23 (2001), no. 2, 97–115.

[26]  Viola, P.A., & Jones, M.J. (2001). Rapid object detection using a boosted cascade of simple features. Proceedings of the 2001 IEEE Computer Society Conference on Computer Vision and Pattern Recognition. CVPR 2001, 1, I-I.

[27]  Y. Taigman, M. Yang, M. Ranzato and L. Wolf, "DeepFace: Closing the Gap to Human-Level Performance in Face Verification," 2014 IEEE Conference on Computer Vision and Pattern Recognition, Columbus, OH, USA, 2014, pp. 1701-1708, doi: 10.1109/CVPR.2014.220.

[28]  Zhang, L., Verma, B., Tjondronegoro, D., & Chandran, V. (2018). Facial Expression Analysis under Partial Occlusion. *ACM Computing Surveys (CSUR)*, 51, 1 - 49. https://doi.org/10.1145/3158369.

## AUTHORS

**Dr Hubert Johnson** is an Associate Professor of Computer Science in the School of Computing at Montclair State University, specializing in Software Engineering, and Project Management. He has given numerous invited presentations on Software Engineering and Project Management both Nationally and Internationally. His research interest also include Secure Software development and AI-driven Project Management.

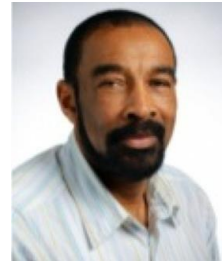

**Swetha Kambham** is a Graduate Assistant in the Computer Science Department in the School of Computing, Montclair State University, New Jersey. Her role includes teaching, tutoring, grading, and research. She was recognized among the top 20% of students for her academic excellence. With a strong background in Software Engineering and Artificial Intelligence, she previously worked as a Software Engineer for two years, gaining expertise in frontend and backend development. Her research interests include AI-driven applications, machine learning, and intelligent systems.

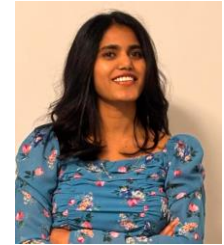

**Sai Prathap Reddy Kambham** is a Full Stack Developer specializing in healthcare technology. He works on software solutions that assist patients facing mental health challenges, especially those struggling with addiction. His work includes developing an advanced pre-intake assessment platform to evaluate patients' initial conditions, helping healthcare professionals provide personalized interventions. Combining expertise in software development and mental health assessments, he contributes to improving patient care and clinical decision-making.

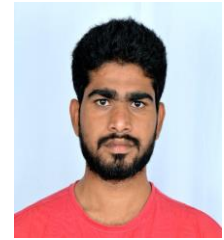